\documentclass[conference]{IEEEtran}
\IEEEoverridecommandlockouts
\usepackage{cite}
\usepackage{amsmath,amssymb,amsfonts}
\usepackage{algorithmic}
\usepackage{graphicx}
\usepackage{textcomp}
\usepackage{xcolor}
\usepackage{booktabs}   
\usepackage{graphicx} 
\usepackage{multirow}  
\usepackage[table]{xcolor} 
\usepackage{array}     
\usepackage[dvipsnames]{xcolor} 
\usepackage{xcolor}
\usepackage{soul}

\usepackage[most]{tcolorbox}
\usepackage{xcolor}
\newtcolorbox{highlight}{
  enhanced,
  sharp corners,
  colback=yellow!40,
  colframe=white,   
  boxrule=0pt,
  breakable,         
  left=2pt, right=2pt, top=2pt, bottom=2pt 
}

\def\BibTeX{{\rm B\kern-.05em{\sc i\kern-.025em b}\kern-.08em
    T\kern-.1667em\lower.7ex\hbox{E}\kern-.125emX}}
\begin{document}

\title{Learning a Semantic Calibration Network for Open-Vocabulary Semantic Segmentation\\
}

\author{\IEEEauthorblockN{Yang Sun$^{1,2}$, Tao Wang$^{2}$, Anastasia Ioannou$^{3}$, Ge Xu$^{2}$}
\IEEEauthorblockA{$^{1}$College of Computer and Data Science, Fuzhou University, Fuzhou, China.\\
$^{2}$School of Computer and Big Data, Minjiang University, Fuzhou, China.\\
$^{3}$Department of Computer Science and Engineering, European University Cyprus, Nicosia, Cyprus.}}

\maketitle

\begin{abstract}
Semantic image segmentation assigns a predefined category label to each pixel, has achieved significant progress lately. Open-Vocabulary Segmentation (OVS) extends the segmentation task from a fixed set to an open set, enabling the identification and segmentation of novel concepts based on arbitrary text inputs, such as category names or descriptions. In this paper, we propose a novel Semantic Calibration Network (SCN) for open-vocabulary semantic segmentation. Different from prior approaches that focus on feature aggregation or simple fine-tuning of pre-trained models, SCN refines the mask classification process by explicitly modeling the semantic correlations between classes, aiming to enhance the model's discriminative power while effectively preserving the generalization abilities of the pre-trained CLIP model. Specifically, SCN comprises two core components: Class Disambiguation (CD) and Logits Fusion (LF). First, a cross-attention mechanism is utilized to transform the text embeddings into visually aware pseudo-text embeddings, in order to derive an enhanced similarity score that complements the original mask-text similarity score. Subsequently, the Class Disambiguation module captures implicit inter-class dependencies through a residual architecture to effectively resolve semantic ambiguities. Finally, the Logits Fusion module dynamically integrates multifaceted semantic evidence to ensure that the model achieves a robust semantic consensus while maintaining CLIP's inherent generalization capability. Comprehensive experimental results on mainstream benchmarks demonstrate that the proposed method achieves significant performance improvements compared to state-of-the-art algorithms.
\end{abstract}

\section{Introduction}
For decades, semantic image segmentation has been a fundamental task in computer vision due to its central role in pixel-level scene understanding and high-level visual reasoning. Despite the success of numerous approaches~\cite{5,6,7,8,9,10,11}, however, traditional segmentation methods cannot localize categories unseen during training, limiting their applicability in open-world scenarios. Open-vocabulary segmentation (OVS) expands the label set in image segmentation from a fixed set to an open set, enabling the segmentation of categories not seen during training. Importantly, open-vocabulary segmentation enables segmentation based on arbitrary text inputs, e.g., categories and descriptions. Recent research often tackles open-vocabulary segmentation by leveraging pre-trained vision-language models, e.g., CLIP~\cite{clip} and ALIGN~\cite{align}, which are pre-trained on large-scale image-text pairs and have strong zero-shot recognition capabilities.

Inspired by this, mainstream OVS methods~\cite{openseg,zegformer,ovseg,odise,fcclip,maft,maft+,san,deop,gba,maskadapter,maskclip+,ebseg} adopt a ``segment-then-recognize" paradigm: it first utilizes a segmentation model to generate class-agnostic mask proposals, and subsequently inputs them into a pre-trained CLIP encoder to perform mask-level visual-language alignment. The objective is to recognize open-vocabulary concepts by combining the generalization abilities of CLIP with the structural priors of the segmentation model. Although significant progress has been made under this paradigm, the image-text alignment abilities of CLIP focuses on global semantics, creating a natural granularity gap with the fine-grained pixel-level annotations required for semantic segmentation.

\begin{figure*}[t!] %
  \centering
  \includegraphics[width=6.5in, page=1]{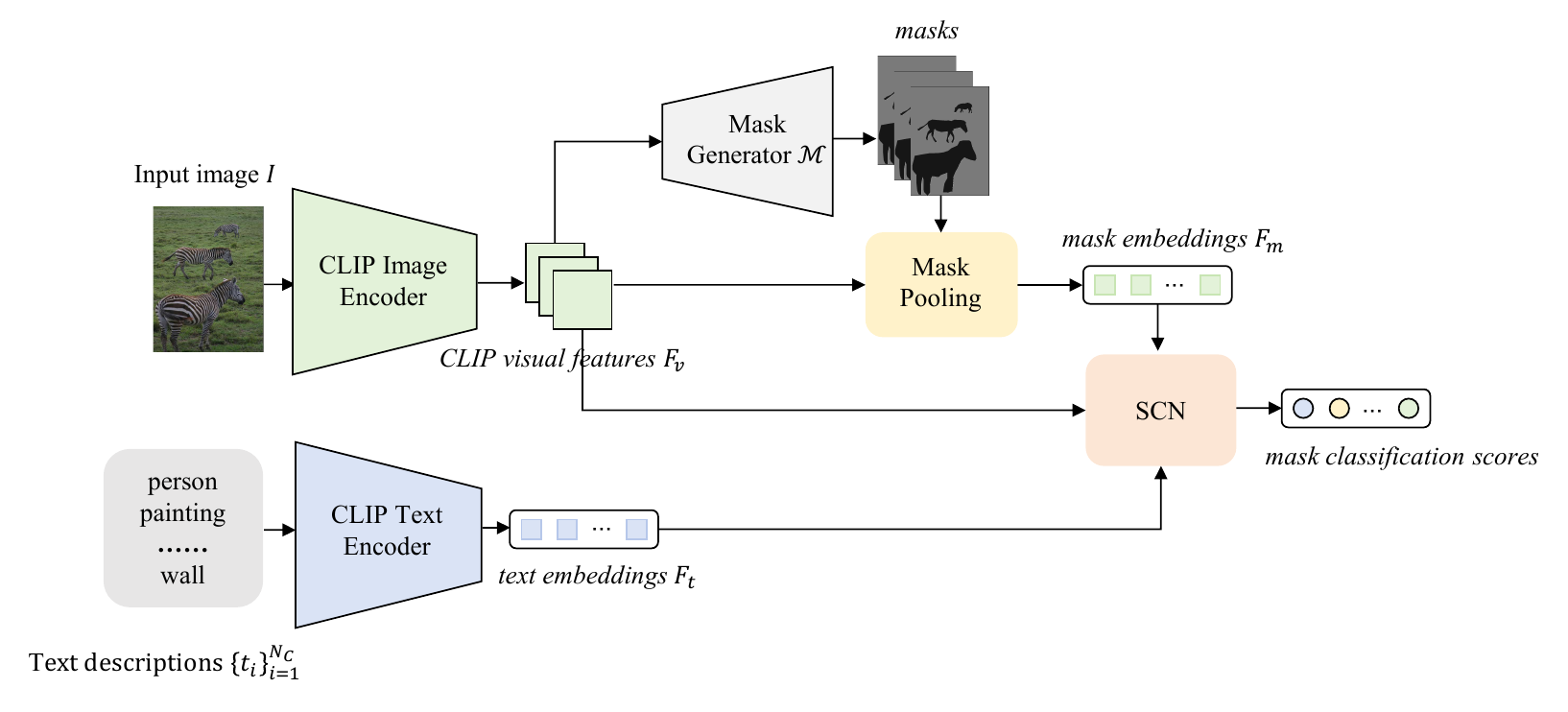} 
  \caption{\textbf{Overview of the proposed framework.} Our model takes an image $\mathcal{I}$ and a set of text descriptions $\{t_i\}_{i=1}^{N_C}$ as input. The vision and language encoders extract visual features $F_v$ and text features $F_t$, respectively. A mask generator $\mathcal{M}$ produces $N$ binary masks $M$ from $F_v$, which are then combined via mask pooling to obtain pooled features $F_m$. Finally, the Semantic Calibration Network (SCN) integrates $F_v$, $F_t$, and $F_m$ to compute similarity scores for the final segmentation map.}
  \label{fig:overpipeline}
\end{figure*}

Consequently, fine-tuning CLIP to adapt it for segmentation tasks has become a research focus. Recent studies (e.g.,\cite{maskclip+}) indicate that the performance bottleneck of such methods no longer lies in the quality of mask generation, but rather in the subsequent mask classification process. Specifically, semantic confusion easily occurs during the classification stage, which severely limits the final segmentation accuracy. Based on these observations, we propose a Semantic Calibration Network, which aims to refine feature representations by explicitly modeling semantic correlations, thereby significantly enhancing the model's discriminative capability on semantically ambiguous categories. Specifically, the Class Disambiguation module captures implicit inter-class dependencies through a residual architecture to effectively resolve semantic ambiguities, and the Logits Fusion module dynamically integrates multi-faceted semantic evidence to ensure that the model achieves a robust semantic consensus while maintaining CLIP’s inherent generalization capability. Fig.~\ref{fig:overpipeline} shows a high-level overview of the proposed method. The primary contributions of this work can be summarized as follows:
\begin{itemize}
\vspace{-1mm}
    \item We propose a Semantic Calibration Network (SCN), which refines feature representations by explicitly modeling semantic correlations, therefore boosting the model's discriminative capability while preserving the zero-shot generalization of the pre-trained CLIP model.
    \item We design two core components to refine feature representations: the Class Disambiguation (CD) module, which resolves semantic ambiguities by capturing implicit inter-class dependencies, and the Logits Fusion (LF) module, which performs a dynamic integration of original and refined similarity scores to achieve a semantic consensus.
    \item We achieve state-of-the-art performance on ADE20K~\cite{ADE} and PASCAL-Context~\cite{pascalcontext} for open-vocabulary segmentation, outperforming existing methods.
\end{itemize}


\section{Related Work}

Open-vocabulary segmentation aims to segment images
based on arbitrary textual descriptions rather than predefined classes. Existing open-vocabulary segmentation methods can be broadly categorized into two streams based on their visual feature organization: pixel-based and mask-based methods. Both approaches aim to exploit pre-trained vision-language models such as CLIP~\cite{clip} for segmentation tasks, focusing on pixel-level cost modeling and region-level feature aggregation, respectively.

\noindent \textbf{Pixel-based methods.} These methods do not involve explicit mask generation, and construct coarse-grained similarity cost maps directly between CLIP’s pixel-level features and text embeddings, followed by multi-stage refinements. For example, SED~\cite{sed} employs a ConvNeXt~\cite{convnext} structure to generate cost maps and performs hierarchical aggregation with multi-layer visual features. CAT-Seg~\cite{catseg} focuses on matching cost refinement, utilizing spatial and class aggregation stages to iteratively converge initial costs into dense semantic maps. Building on this, ESC-Net~\cite{escnet} integrates correlation pseudo-prompts with SAM's regional aggregation capabilities to enhance early-stage semantic structure. In general, pixel-based methods avoid the instability of mask generation and directly model pixel-level similarities, making them well-suited for capturing fine-grained boundaries and complex structures.

\noindent \textbf{Mask-based methods.} These methods adopt a ``segment-then-recognize" paradigm: first generating category-agnostic binary masks, and then aggregating visual features within these regions for classification via cosine similarity with text embeddings. 
The seminal work OpenSeg \cite{openseg} first employs mask pooling to aggregate region features for mask classification. Zegformer~\cite{zegformer} proposes a two-stage framework that generates class-agnostic masks before utilizing CLIP for classification, establishing the standard pipeline for this approach. Subsequently, OVSeg~\cite{ovseg} introduces a mask-adaptive fine-tuning strategy to bridge the domain gap between pre-trained CLIP and non-rectangular mask regions, significantly improving the accuracy of region classification. Furthermore, ODISE~\cite{odise} leverages diffusion models to generate high-quality semantic masks, while FC-CLIP~\cite{fcclip} streamlines the architecture by employing a shared frozen convolutional backbone to handle both mask generation and class prediction simultaneously.

In addition, various enhancement strategies have been proposed for feature alignment. For instance, MAFT~\cite{maft} fine-tunes the CLIP image encoder while employing a self-distillation loss to prevent catastrophic forgetting, and MAFTP~\cite{maft+} further introduces content-dependent transfer to refine text representations. SAN~\cite{san} utilizes a side adapter network to bridge the gap between mask predictions and CLIP features. DeOP~\cite{deop} improves region classification quality by suppressing noise interaction between pixels and introducing classification anchors. Additionally, EBSeg~\cite{ebseg} mitigates overfitting on training classes by dynamically fusing generalizable and discriminative features; GBA~\cite{gba} enhances region features in the frequency domain; while MaskAdapter~\cite{maskadapter} and MaskCLIP++~\cite{maskclip+} focus on more robust mask feature aggregation mechanisms and consistency alignment constraints. Overall, mask-based methods primarily focus on maintaining precise cross-modal alignment at the mask-level feature level.

Despite these advancements, a fundamental spatial granularity discrepancy issue persists between CLIP's global image-text pre-training and the fine-grained boundary delineation requirements for semantic segmentation. We argue that simply fine-tuning mask or text features often leads to overfitting on seen classes, undermining the model's inherent zero-shot generalization capability. In particular, MaskCLIP++~\cite{maskclip+} suggests that the key to maintaining generalization lies in preserving the relative order of original similarity scores. Therefore, we propose a Semantic Calibration Network in this work. Unlike previous methods focused solely on feature aggregation, SCN is designed to explicitly model semantic correlations to enhance feature representations. By calibrating the feature distribution, SCN boosts the model's discriminative capability on semantically ambiguous categories while preserving the zero-shot generalization abilities of the pre-trained CLIP backbone. To the best of our knowledge, this is the first work in open-vocabulary semantic segmentation that explicitly models inter-class dependencies directly in the similarity~(logit) space for semantic calibration.

\section{Proposed Method}

In this section, we describe the open vocabulary segmentation problem and our proposed method in detail. We first revisit the problem formulation and an overview to the mask-based method in Sec.~\ref{sec:prelim}. This is followed by details of the proposed Semantic Calibration Network in Sec.~\ref{sec:scnmethod} that includes an enhanced similarity score, a Class Disambiguation (CD) module, and a Logits Fusion (LF) module.

\subsection{Preliminaries}\label{AA}
\label{sec:prelim}
Open-vocabulary segmentation aims to segment a given image $\mathcal{I} \in \mathbb{R}^{H \times W \times 3}$ into a set of masks with associated semantic labels: 
\begin{equation}
    \{y_i\}_{i=1}^K=\{(m_i, c_i)\}_{i=1}^K,
\end{equation}
where $K$ is the number of ground truth masks in the image, $m_i \in \{0, 1\}^{H \times W}$ and $c_i \in \{0,1\}^{|\mathcal{C}_{train}|}$ denote the $i$-th binary ground-truth mask and
its corresponding one-hot encoded class label, respectively. Here, $|\mathcal{C}_{train}|$ denotes the number of classes in the training set. During training, a fixed set of classes $\mathcal{C}_{train}$ is used, while during inference, another set of classes $\mathcal{C}_{test}$ is employed. 
More specifically, $\mathcal{C}_{test}$ usually contains entirely new classes that do not appear in $\mathcal{C}_{train}$, i.e., $\mathcal{C}_{test}\cap\mathcal{C}_{train} = \emptyset$.

Mask-based OVS methods first generate category-agnostic mask proposals via a mask generator, and subsequently classify these proposals using a vision language model such as CLIP to obtain the final segmentation results. Concretely, the first stage consists of a category-agnostic mask generator $\mathcal{M}$, parameterized by $\Theta_\mathcal{M}$. Given an input image $\mathcal{I}$, it generates $N$ mask proposals: 
\begin{equation}\{\widehat{m}_{i}\}_{i=1}^{N}=\mathcal{M}\big(\mathcal{I};\Theta_{\mathcal{M}}\big).\end{equation}

\noindent where $\widehat{m}_i\in\mathbb{R}^{H\times W}$ is the $i$-th mask proposal. In the next stage, the CLIP adapter $\mathcal{P}$ uses the mask proposals $\{\widehat{m}_i\}_{i=1}^N$ to provide spatial guidance when extracting visual features for image $\mathcal{I}$. Mask proposals can guide the model to precisely attend to specific regions, in order to obtain mask-level embeddings. Subsequently, the adapter $\mathcal{P}$ achieves cross-modal alignment by computing the similarity between these mask-level embeddings and the class-specific text embeddings, so as to classify each mask proposal. More formally, this process can be written as:

\begin{equation}\{\widehat{c}_i\}_{i=1}^N=\mathcal{P}\big(\mathcal{I},\{\widehat{m}_i\}_{i=1}^N\big)\end{equation}
where $\widehat{c}_i\in\mathbb{R}^{|\mathcal{C}_{test}|}$ denotes the predicted class probabilities of the $N$ mask proposals for each category, and $|\mathcal{C}_{test}|$ denotes the number of classes in the test set.

Following the mask-based method above, the overall architecture of our model is illustrated in Fig.~\ref{fig:overpipeline}. The framework comprises a pre-trained CLIP vision encoder, a text encoder, a class-agnostic mask generator, and the proposed Semantic Calibration Network. More specifically, given an input image $\mathcal{I}$ and a set of text descriptions $\{t_i\}_{i=1}^{N_C}$, where $N_C$ denotes the cardinality of the text description set, i.e., $N_C=|\mathcal{C}_{train}|$ during training and $N_C=|\mathcal{C}_{test}|$ during test, the vision and text encoders first extract the visual features $F_v\in\mathbb{R}^{C\times H\times W}$ and text embeddings $F_t\in\mathbb{R}^{C\times {N_C}}$, respectively. Subsequently, the mask generator $\mathcal{M}$ yields $N$ binary mask proposals $\{\widehat{m}_i\}_{i=1}^N$. A mask pooling module then aggregates ${F}_{v}$ under the guidance of $\mathcal{M}$ to derive the pooled mask-level CLIP features ${F}_{m}\in\mathbb{R}^{C\times N}$. In vanilla mask-based methods, the pairwise similarity score between $F_m$ and $F_t$, which we later denote as ${S}\in\mathbb{R}^{N\times {N_C}}$, is used to classify the mask proposals. In contrast, ${F}_{m}$, ${F}_{v}$, and ${F}_{t}$ are fed into the SCN in our work to compute the refined similarity score ${Q}\in\mathbb{R}^{N\times {N_C}}$ which is used for mask classification. We will discuss detailed steps in the next section.

\begin{figure*}[!t] %
  \centering
  \includegraphics[width=5.5in, page=1]{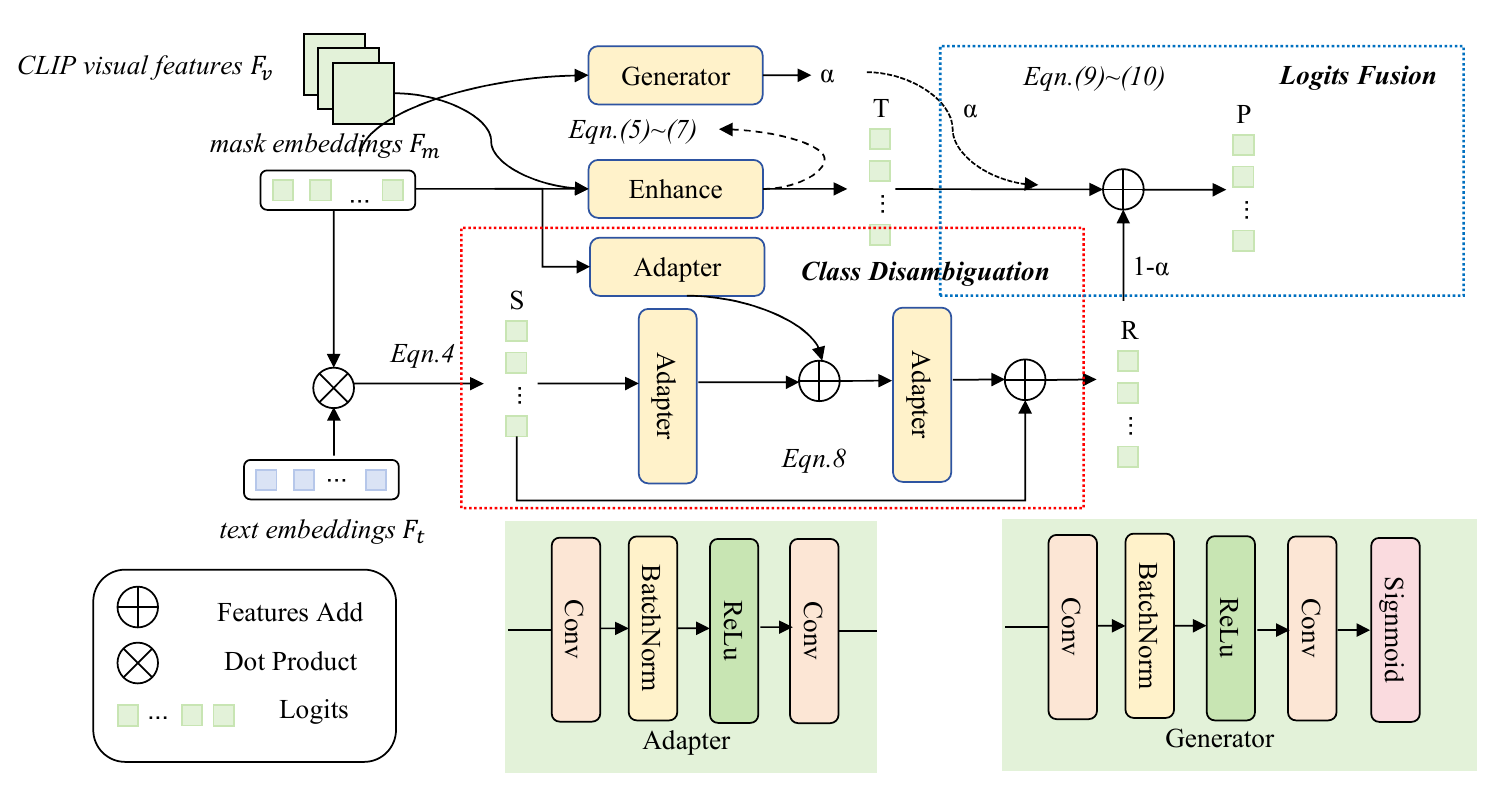} 
  \caption{\textbf{Architecture of the Semantic Calibration Network.} SCN derives dual similarity maps and integrates Class Disambiguation (CD) and Logits Fusion (LF) modules based on a lightweight generator to strengthen discriminative power by explicitly modeling semantic correlations while preserving CLIP's generalization abilities.}
  \label{fig:scm}
\end{figure*}

\subsection{Semantic Calibration Network}\label{AA}
\label{sec:scnmethod}

In this section, we move on to describe the proposed Semantic Calibration Network in detail. One of the key challenges for mask-based OVS methods is a representation mismatch. Specifically, CLIP is pre-trained on image-text pairs, so its learned features primarily support global image-level alignment. In contrast, semantic segmentation requires fine-grained pixel-level classification guided by dense annotations. Between them, there is a clear discrepancy in terms of spatial granularity and task objectives. Therefore, how to effectively enhance CLIP representations to boost its performance in dense prediction tasks is a primary focus of research. In particular, simply fine-tuning mask features or text features will likely lead to overfitting on seen classes, undermining CLIP's zero-shot generalization capability. We note that, MaskCLIP++~\cite{maskclip+} identified that the key to maintaining CLIP's generalization capability lies in keeping the relative order of its original similarity scores unchanged. Inspired by this observation, we propose the Semantic Calibration Network. SCN is designed to enhance feature representations by explicitly modeling semantic correlations, thereby boosting the model's discriminative capability on semantically ambiguous classes while preserving the zero-shot generalization abilities of the pre-trained CLIP model. As illustrated in Fig.~\ref{fig:scm}, SCN consists of two core components: Class Disambiguation (CD) and Logits Fusion (LF). However, in order to provide these components with multi-faceted semantic evidence, we first derive an additional similarity score in addition to the standard mask-text similarity ${S}\in\mathbb{R}^{N\times {N_C}}$. Specifically, an enhanced similarity score ${T}\in\mathbb{R}^{N\times {N_C}}$ is computed by integrating multi-modal inputs, including mask, text, and visual features. Mathematically, the computations for ${S}$ and ${T}$ are formulated as follows:

\noindent \textbf{Standard similarity score.} The standard similarity score ${S} \in \mathbb{R}^{N \times N_C}$ is obtained via the dot-product between mask embeddings ${F}_m$ and text features ${F}_t$:
\begin{equation}S=F_m^{\mathrm{T}}F_t\end{equation}

\noindent \textbf{Enhanced similarity score.} The enhanced similarity score ${T} \in \mathbb{R}^{N \times N_C}$ aims to inject visual features into the text embeddings through cross-modal interaction. This allows the resulting similarity to dynamically attend to different local regions adaptively. Specifically, we instantiate a cross-attention mechanism where text features serve as the query to selectively aggregate pertinent visual cues from visual features. This process effectively transforms the original text embeddings into visually aware pseudo-text embeddings. To augment the discriminative power while preserving the zero-shot generalization of the pre-trained CLIP model, a residual architecture is used to preserve the original text representation while enabling incremental updates. Concretely, this can be written as:

\begin{equation}P_t=\mathrm{Attention}(q(F_t),k(F_v),v(F_v))\end{equation}
\begin{equation}F_t^{^{\prime}}=\mathrm{Norm}(\mathrm{F}_t+P_t)\end{equation}
\begin{equation}T=F_m^{\mathrm{T}}F_t^{\prime}\end{equation}

\noindent where $\mathrm{Attention}(\cdot)$ denotes the scaled dot-product attention mechanism, and $\{q, k, v\}$ represent the input linear projection functions within the attention layer. $\mathrm{Norm}(\cdot)$ denotes the Layer Normalization operation, while $T$ denotes the enhanced similarity score derived via the dot-product operation between mask and refined text embeddings.

Next, building upon these representations, we propose the following modules: (1) Class Disambiguation (CD) that takes mask embeddings as priors and employs a residual architecture to capture inter-class dependencies. By adaptively adjusting similarity scores $S$ based on mask-specific features $F_m$ and the similarity score $S$ itself, this process encodes second-order mask-specific dependencies and effectively resolves semantic ambiguities, yielding a refined similarity score. (2) Logits Fusion (LF) that performs a dynamic integration of the standard similarity score after CD, and the enhanced similarity score we obtained in the previous step. Notably, the adaptive fusion weights are derived from a lightweight generator module, ensuring an optimal balance between multi-source semantic evidence. Specific steps are as follows.


\noindent \textbf{Class Disambiguation.} To resolve the issue of semantic ambiguity, we propose a Class Disambiguation module based on a residual architecture. The objective is to learn from and eliminate ambiguities within the relatively coarse similarity maps ${S} \in \mathbb{R}^{N \times N_C}$, thereby yielding a more refined similarity map ${R} \in \mathbb{R}^{N \times N_C}$.
Specifically, the module first explores latent inter-class dependencies from ${S} \in \mathbb{R}^{N \times N_C}$ via adapter $\mathcal{A}_1$, while simultaneously utilizing adapter $\mathcal{A}_2$ to capture prior information related to semantic ambiguity from mask features ${F}_{m}\in\mathbb{R}^{N\times C}$. Subsequently, the outputs of both adapters are fed into adapter $\mathcal{A}_3$ to jointly learn the correlation between the similarity map $S$ and mask features $F_m$, thereby eliminating semantic ambiguity. The detailed process is as follows:
\begin{equation}R=S+\mathcal{A}_3(\mathcal{A}_1(S)+\mathcal{A}_2(F_m))\end{equation}

\noindent We operate directly on similarity logits instead of feature embeddings in order to preserve CLIP’s original embedding geometry and relative similarity ordering, which is crucial for maintaining zero-
shot generalization. See Fig.~\ref{fig:scm} for the network structure of the adapters $\mathcal{A}_1$, $\mathcal{A}_2$ and $\mathcal{A}_3$.

\begin{table}[t!]
\centering
\caption{Layer parameters of the semantic calibration Network.}
\label{tab:parameters}
\begin{tabular}{ccc}
\toprule
 & layer parameters & output size \\ \midrule
\multicolumn{3}{l}{adapter $\mathcal{A}_1$} \\ \midrule
$fc1$ & $1 \to 512$ & ${N\times N_c\times 512 } $ \\
$conv1$ & $1 \times 1 \times 640$, stride 1 & ${N\times N_c\times 640 } $\\
$conv2$ & $1 \times 1 \times 640$, stride 1 & ${N\times N_c\times 640 } $\\ \midrule
\multicolumn{3}{l}{adapter $\mathcal{A}_2$} \\ \midrule
$conv3$ & $1 \times 1 \times 640$, stride 1 & ${N\times N_c\times 640 } $\\
$conv4$ & $1 \times 1 \times 640$, stride 1 & ${N\times N_c\times 640 } $\\ \midrule
\multicolumn{3}{l}{adapter $\mathcal{A}_3$} \\ \midrule
$conv5$ & $1 \times 1 \times 640$, stride 1 & ${N\times N_c\times 640 } $\\
$conv6$ & $1 \times 1 \times 512$, stride 1 & ${N\times N_c\times 512 } $\\ 
$fc2$ & $512 \to 1$ & ${N\times N_c\times 1 } $\\ \midrule
\multicolumn{3}{l}{Generator} \\ \midrule
$conv7$ & $1 \times 1 \times 640$, stride 1 & ${N\times N_c\times 640 } $\\
$conv8$ & $1 \times 1 \times 1$, stride 1 & ${N\times N_c\times 1 } $\\
\bottomrule

\end{tabular}
\end{table}

\noindent \textbf{Logits Fusion.} We propose a Logits Fusion module, which performs a weighted fusion of the refined similarity map obtained from Class Disambiguation and the enhanced similarity map from the previous step. Specifically, we design a fusion weight generator that generates fusion weights from mask features to perform weighted fusion on ${R} \in \mathbb{R}^{N \times N_C}$ and ${T} \in \mathbb{R}^{N \times N_C}$, in order to obtain the final similarity score ${Q} \in \mathbb{R}^{N \times N_C}$, which is used for mask classification. The detailed process is as follows:
\begin{equation}\alpha=\mathrm{Generator}(F_m)\end{equation}
\begin{equation}Q=\alpha T+(1-\alpha)R\end{equation}

\noindent We refer readers to Fig.~\ref{fig:scm} for the network structure of the generator. With the aid of the Logits Fusion module, the model adaptively models and integrates complementary semantic predictions with mask feature representations, thereby achieving the dynamic calibration of cross-category feature correlations. This process effectively enhances the semantic consistency and discriminability of the final similarity map, while simultaneously preserving the model's generalization capability in open-domain scenarios.

Table~\ref{tab:parameters} shows the layer parameters for adapters $\mathcal{A}_1$, $\mathcal{A}_2$, $\mathcal{A}_3$ and the generator in SCN. In terms of model learning, the parameters required in the enhanced similarity scores, the adapters and the generator can be simultaneously learned with the rest of the network, resulting in a standard end-to-end learning procedure as other mask-based OVS methods, using the standard cross-entropy loss.

\section{Experiments}

\subsection{Datasets and Evaluation Metric}
Following the common practice in open-vocabulary segmentation, we use COCO-Stuff \cite{coco} as our training foundation. This dataset provides a rich set of approximately 118k images covering 171 semantic classes. To rigorously evaluate the performance and generalization capability, we evaluate our model on ADE20K \cite{ADE}, PASCAL
VOC \cite{pascalvoc}, and PASCAL-Context \cite{pascalcontext} datasets. Details for
each dataset are as follows:
\begin{itemize}
    \item ADE20K \cite{ADE}: A comprehensive dataset consisting of 20,000 training and 2,000 validation samples. We report results on both the A-150 (comprising 150 frequent categories) and the A-847 (a more challenging split with 847 categories) subsets.
    \item PASCAL-VOC \cite{pascalvoc}: Also known as PAS-20 in this context, it includes about 1,500 images for training and validation respectively, across 20 object classes.
    \item PASCAL-Context \cite{pascalcontext}: An expanded version of PASCAL VOC. We evaluate on two standard protocols: PC-59 (59 categories) and the fine-grained PC-459 (459 categories).
\end{itemize}

To quantitatively evaluate the performance, we follow existing traditional open-vocabulary semantic segmentation. Semantic segmentation results are evaluated with mean Intersection over Union (mIoU).

\subsection{Implementation Details}
Our model is implemented based on the Detectron2 framework. We use CLIP with a ConvNeXt-B backbone as our vision language model, which extracts visual features with a dimension of 640. For optimization, we use the AdamW optimizer with an initial learning rate of $1\times10^{-4}$ and a weight decay of 0.05, coupled with a multi-step learning rate decay schedule. The model is trained on the COCO-Stuff dataset for 5 epochs with a total batch size of 12, utilizing two NVIDIA RTX A6000 GPUs. During training, we apply data augmentation techniques including random horizontal flipping and a crop size of $1024\times1024$. For evaluation, the image resolution is set to $896\times896$ to maintain consistency across benchmarks.

\begin{table*}[t]
\centering
\small 
\caption{Open-vocabulary semantic segmentation performance. The best results are indicated in bold. We use mIoU as the evaluation
metric.*~denotes that the results are re-evaluated.}
\label{tab:performance}
\resizebox{\textwidth}{!}{%
\begin{tabular}{lll|ccccc}
\toprule
\textbf{Method} & \textbf{VLM} & \textbf{Training Dataset} & \textbf{A-150} & \textbf{A-847} & \textbf{PC-59} & \textbf{PC-459} & \textbf{PAS-20} \\ \midrule
ZegFormer\cite{zegformer}  & CLIP ViT-B/16 & COCO-Stuff & 18.0 & 5.6 & 45.5 & 10.4 & 89.5 \\
DeOP\cite{deop}  & CLIP ViT-B/16 & COCO-Stuff & 22.9 & 7.1 & 48.8 & 9.4 & 91.7 \\
OvSeg\cite{ovseg}  & CLIP ViT-B/16 & COCO-Stuff & 24.8 & 7.1 & 53.3 & 11.0 & 92.6 \\
SAN\cite{san}  & CLIP ViT-B/16 & COCO-Stuff & 27.5 & 10.1 & 53.8 & 12.6 & 94.0 \\
OpenSeg\cite{openseg}  & ALIGN & COCO Panoptic+Loc. Narr. & 28.6 & 8.8 & 48.2 & 12.2 & 72.2 \\
EBSeg\cite{ebseg}  & CLIP ViT-B/16 & COCO-Stuff & 30.0 & 11.7 & 56.7 & 17.3 & 94.6 \\
SED\cite{sed} & CLIP ConvNeXt-B & COCO-Stuff & 31.6 & 11.4 & 57.3 & 18.6 & 94.4 \\
CAT-Seg\cite{catseg}  & CLIP ViT-B/16 & COCO-Stuff & 31.8 & 12.0 & 57.5 & \textbf{19.0} & 94.6 \\
MAFTP*\cite{maft+}  & CLIP ConvNeXt-B & COCO-Stuff & 34.5 & 13.8 & 57.5 & 18.5 & \textbf{95.5} \\
MAFTP w/ MaskAdapter~(Baseline)\cite{maskadapter} & CLIP ConvNeXt-B & COCO-Stuff & 35.6 & 14.2 &58.4 & 17.9 & 95.1 \\ 

SCN~(Ours) & CLIP ConvNeXt-B & COCO-Stuff & \textbf{36.2} & \textbf{14.7} & \textbf{58.6} & 18.8 & 95.1 \\
\bottomrule
\end{tabular}
} 
\end{table*}

\begin{table}[h]
\centering
\caption{Ablation study of our model. BL: the baseline model; CD: the class disambiguation module; LF: the logits fusion module.}
\begin{tabular}{ccc|ccccc}
\toprule
BL & CD & LF & A-150 & A-847 & PC-59 & PC-459 & PAS-20 \\ \midrule
\checkmark &            &            & 35.6 & 14.2& 58.4 & 17.9 & 95.1
\\
\checkmark & \checkmark &            & 35.9 & 14.5 & 58.5 & 18.5 & 95.1\\
\checkmark & \checkmark & \checkmark & 36.2 &  14.7 & 58.6 &18.8 & 95.1\\ \bottomrule
\end{tabular}
\vspace{8pt}
\label{tab:ablation}
\end{table}

\subsection{Results}

\noindent \textbf{Quantitative Comparison}.
Table~\ref{tab:performance} presents the quantitative results of the proposed framework compared to previous state-of-the-art methods on standard open-vocabulary semantic segmentation benchmarks, following standard evaluation protocols. We categorize the models based on the size of the Vision-Language Model (VLM). For base-scale VLM models, our SCN achieves the best performance on the A-150, A-847, and PC-59 datasets, with mIoU scores of 36.2, 14.7, and 58.6, respectively. Compared to the baseline model, we improve mIoU by 0.6, 0.5, 0.2, and 0.9 on the A-150, A-847, PC-59, and PC-459 datasets, respectively, Performance gains are mainly attributed to the proposed Semantic Calibration Network rather than backbone changes. which fully demonstrates the effectiveness of our method.

\noindent \textbf{Ablation Study}. To evaluate the individual contribution of each component, we conduct a systematic ablation study by incrementally adding them to the baseline. As summarized in Table~\ref{tab:ablation}, BL denotes the Mask-Adapter baseline, while CD and LF represent the proposed Class Disambiguation and Logits Fusion modules, respectively. Starting from the baseline (35.6 mIoU on A-150), incorporating the CD module yields a performance gain of 0.3, reaching 35.9. The integration of the LF module further boosts the performance to a peak of 36.2 mIoU. This consistent improvement demonstrates the effectiveness of both CD and LF modules in enhancing open-vocabulary semantic segmentation.

\begin{figure*}[!t] %
  \centering
  \includegraphics[width=5.5in, page=1]{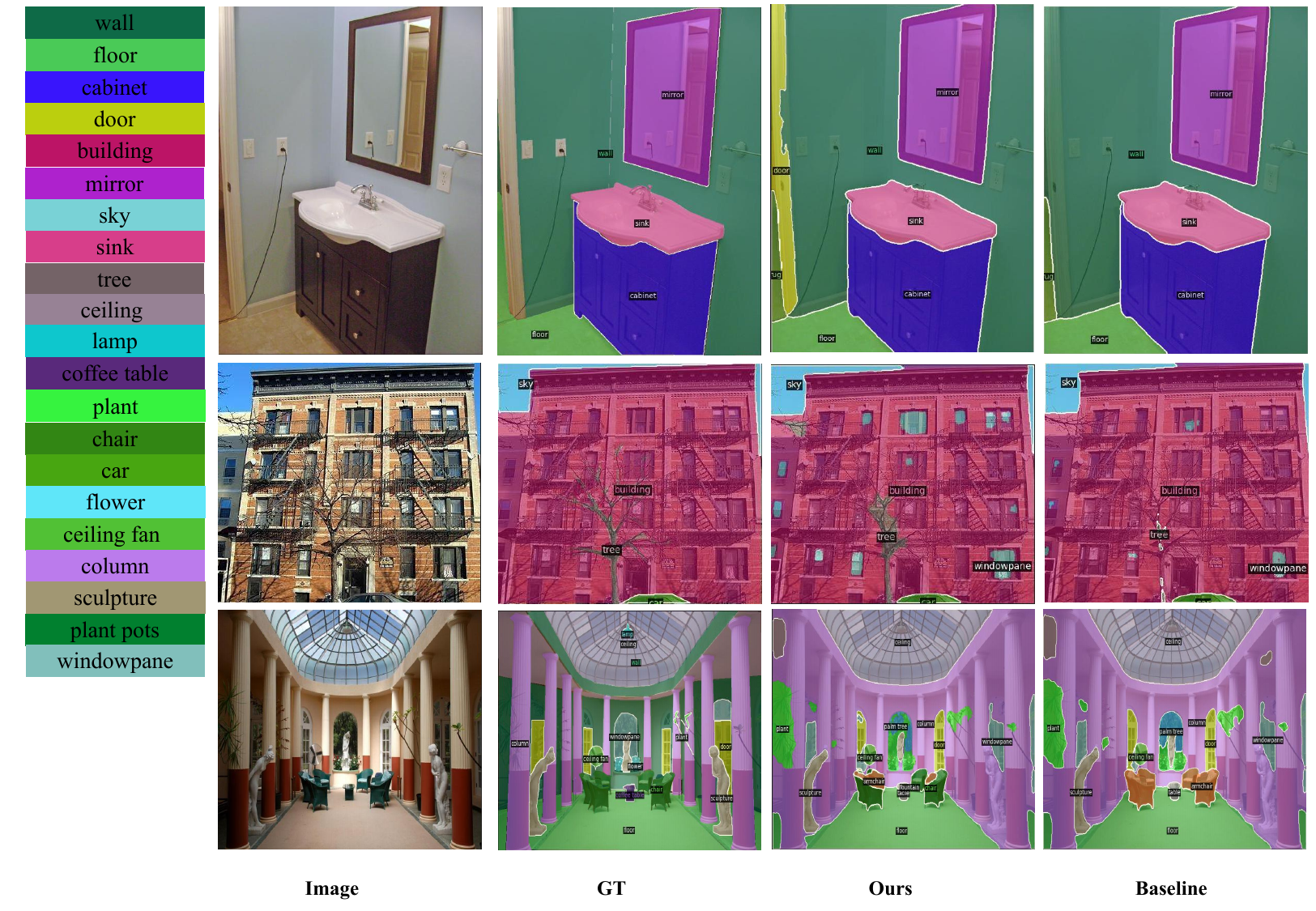} 
  \caption{\textbf{Qualitative results.} The results obtained with our method and the MAFTP with MaskAdapter~\cite{maskadapter} baseline are shown for comparison.}
  \label{fig:image}
\end{figure*}

\noindent \textbf{Qualitative Results.}
We visualize the prediction results on the ADE20K dataset alongside the ground-truth (GT) annotations to qualitatively demonstrate the performance of our model. Some examples are shown in Fig.~\ref{fig:image}. As illustrated in the third row, the baseline model exhibits significant semantic confusion when dealing with visually similar categories; specifically, it misclassifies ``chair'' as ``armchair,'' whereas our model effectively rectifies this issue. In the second row, our model achieves precise segmentation of the ``tree'' region. In addition, the first row shows that our model successfully segments ``door'', Although ``door'' is not annotated in the ground-truth of this sample, the predicted region corresponds to a visually valid door instance, illustrating the open-vocabulary generalization capability of the model.

\section{Conclusion}
In this work, we propose a novel Semantic Calibration Network for open-vocabulary semantic segmentation. By leveraging a cross-attention mechanism, SCN transforms the pre-trained text embeddings into visually aware pseudo-text embeddings, deriving an enhanced similarity score that effectively complements the original mask-text similarity score. Unlike previous methods, our approach explicitly models inter-class dependencies via the Class Disambiguation module, while the Logits Fusion module dynamically integrates multi-faceted semantic evidence to preserve CLIP's inherent generalization abilities. Experimental results show that SCN outperforms current state-of-the-art methods across multiple benchmarks.

\nocite{*}
\bibliographystyle{IEEEtran}
\bibliography{refs}

\end{document}